\PassOptionsToPackage{table}{xcolor}
\documentclass[runningheads]{llncs}
\usepackage[T1]{fontenc}
\usepackage{booktabs}
\usepackage[skins]{tcolorbox}
\usepackage{soul}
\tcbuselibrary{breakable}

\usepackage{subcaption}
\usepackage{xcolor}
\usepackage{colortbl}
\usepackage{graphicx}
\usepackage{multirow}

\definecolor{lightblue}{RGB}{173, 216, 230} 
\definecolor{lightpurple}{RGB}{221, 160, 221} 
\definecolor{brightblue}{RGB}{135, 206, 250} 

\begin{document}
\title{Psychology-Driven Enhancement of Humour Translation}
%
%
\author{Yuchen Su\inst{1}\orcidID{0009-0007-5287-4163} \and
Yonghua Zhu\inst{2}\orcidID{0000-0003-3339-8855} \and
Yang Chen\inst{1}\orcidID{0000-0002-1148-3920} \and
Diana Benavides Prado\inst{3}\orcidID{0000-0003-1137-2822} \and
Michael Witbrock\inst{1}\orcidID{0000-0002-7554-0971}}
\authorrunning{F. Author et al.}
%
\institute{School of Computer Science, University of Auckland, New Zealand \and Singapore University of Technology and Design \and School of Electronic Engineering and Computer Science, Queen Mary University of London \\
\email{\{ysu132\}@aucklanduni.ac.nz}}
\maketitle              
\begin{abstract}
Humour translation plays a vital role as a bridge between different cultures, fostering understanding and communication. Although most existing Large Language Models (LLMs) are capable of general translation tasks, these models still struggle with humour translation, which is especially reflected through linguistic interference and lacking humour in translated text. In this paper, we propose a psychology-inspired Humour Decomposition Mechanism (HDM) that utilises Chain-of-Thought (CoT) to imitate the ability of the human thought process, stimulating LLMs to optimise the readability of translated humorous texts. Moreover, we integrate humour theory in HDM to further enhance the humorous elements in the translated text. Our automatic evaluation experiments on open-source humour datasets demonstrate that our method significantly improves the quality of humour translation, yielding average gains of 7.75\% in humour, 2.81\% in fluency, and 6.13\% in coherence of the generated text. 

\end{abstract}
\section{Introduction}
Humour plays an important role in human interaction. Humour studies can actually gain greater insight into the linguistic, social and psychological factors of humour \cite{zabalbeascoa2005humor}. A comprehensive understanding of humour necessitates a deep grasp of both semantic information and cultural background \cite{chen2024talk}. Effective humour translation serves as a bridge across cultural divides, facilitating communication and fostering cross-cultural understanding \cite{vandaele2016translating}. Some studies \cite{pym2023exploring} mention that the humour translation research can enhance the understanding of language transfer and the process of meaning reconstruction, while enriching the translation theories, especially for dynamic equivalence and functionalist translation strategies. Moreover, an effective humour translation strategy can accurately convey its intended humorous effect in the target language \cite{zabalbeascoa2005humor} and contribute to advancements in general translation research.

\begin{figure}
\centering
\includegraphics[width=0.8\textwidth]{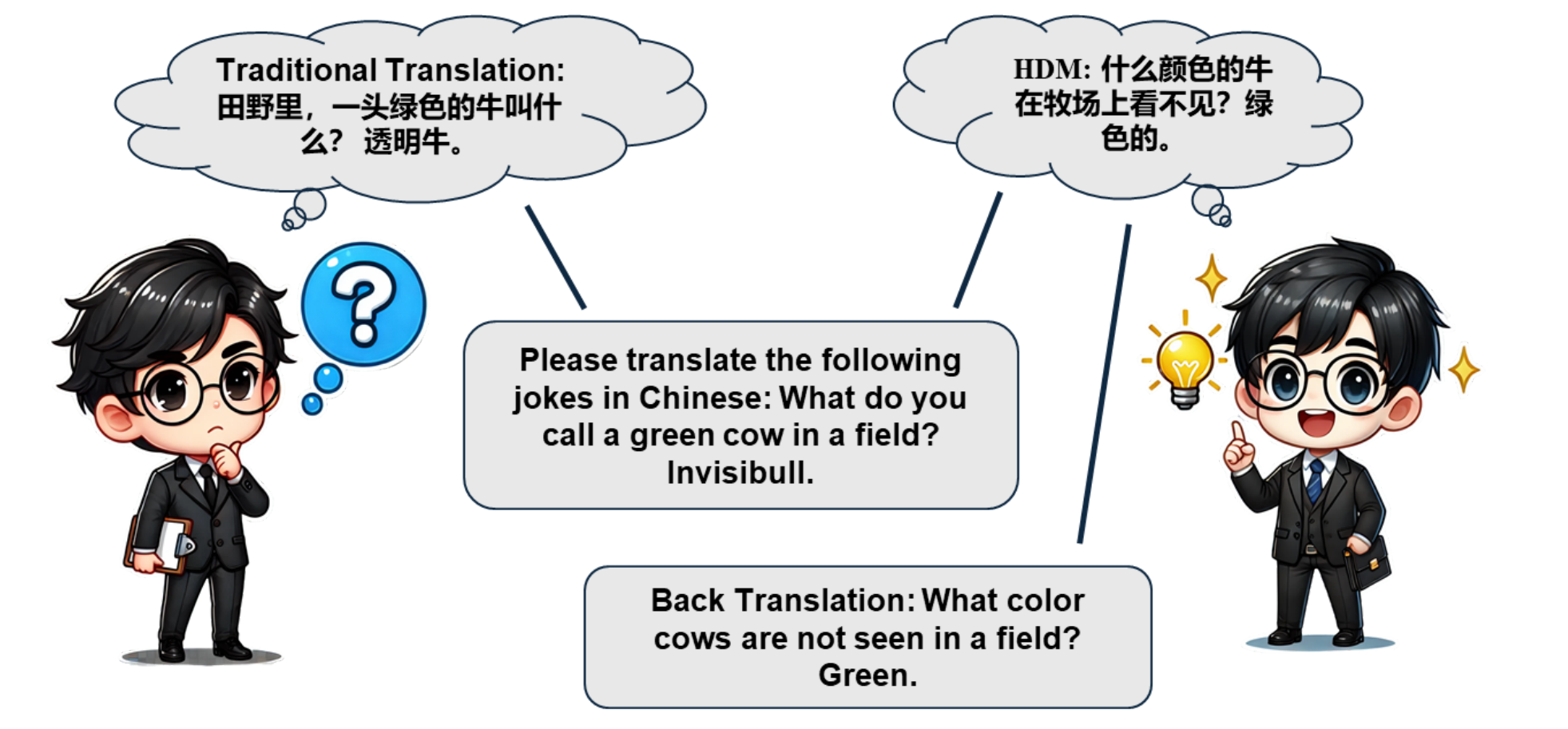}
\caption{An example of humour translation from English to Chinese.}
\label{traditional}
\end{figure}

There are two fundamental approaches to general translation \cite{nida1964toward}: formal equivalence, which prioritizes literal translation, and dynamic equivalence, which focuses on emotional or contextual translation. However, the majority of existing studies focus on literal translation, with limited research exploring emotional translation, particularly in the context of humour. Chen et al. \cite{chen2022towards} use cross-language transfer to enable zero-shot neural machine translation and Wang et al. \cite{wang2022efficient} explore a more efficient KNN-MT for translation. With the advent of large language models (LLMs) such as ChatGPT\footnote{https://chat.openai.com/chat} and GPT-4 \cite{achiam2023gpt}, translation has become a prominent domain where LLMs demonstrate remarkable capacity and competence \cite{zhang2023prompting,he2024exploring}. However, these models still lack proficiency in humour translation in some cases. In Fig \ref{traditional}, for example, the punchline ``Invisibull'' is awkwardly rendered in the Chinese translation, resulting in a stiff and unnatural expression. By altering the linguistic structure and logical order, while preserving the intended meaning, the translation becomes more fluent and natural, as shown on the right side.


Due to linguistic and cultural barriers, humour translation often results in the loss of humour in the translated content \cite{xia2023humor}. The reason is that jokes often rely on extensive knowledge and common sense, and the punchline is usually hidden in the semantics of the sentence, such as cultural context, wordplay, and metaphorical expressions. These elements are challenging to identify and translate accurately \cite{hasan2021humor}, which weakens the humour of the joke in the target language. Additionally, the issue of linguistic interference is a factor in humour translation \cite{hopkinson2007factors}, which is a non-standard version of the target language in the product of translation. Ma and Cheung \cite{ma2020language} indicate that linguistic interference is linked to reduced lexical variety and less cohesive discourse, while the traditional method of translation usually involves merely a linear arrangement of words or phrases \cite{gambier2016translations}, which can result in a lack of fluency and coherence in the translated text. This requires a process that can provide a human thinking process to reconstruct the translated text.

Therefore, to address the challenge of humour translation across different languages, we propose a novel Humour Decomposition Mechanism (HDM) to improve linguistic interference, which introduces a three-step paradigm through the Chain-of-Thoughts (CoT) prompting method \cite{wei2022chain,zhang2022automatic} by utilising LLMs: (1) mining intrinsic knowledge related to the joke; (2) translating the intrinsic knowledge text; and (3) constructing a new joke based on the translated content. This method mimics a human thinking process for understanding, translating and generating, which allows reconstructing the translated text. Furthermore, to enhance humour in translated texts, we integrate humour theory into intrinsic knowledge by defining corresponding topics, angles, and punchlines. This approach enables the model to perform humour translations effectively based on the mined knowledge.

We use the \textit{Estimation Metric Based Assessment} (GEMBA) \cite{kocmi2023large}, a type of LLM evaluation, to assess humour, fluency and coherence. Experimental results reveal that our method is demonstrably superior to existing solutions, showing an average improvement of 7.75\% in humour, 2.81\% in fluency, and 6.13\% in coherence from English to Chinese. These findings indicate that the approach effectively mitigates humour loss and linguistic interference. The main contributions of this paper are summarized as follows:
\begin{itemize}
\item We propose an efficient Humour Decomposition Mechanism to guide LLMs to translate jokes, mimicking the human thought process.
\item This work is the first attempt to incorporate the Psychological theory of constructing humour into the Chain-of-Thought process to improve the humour factors. 

\item The extensive experimental results demonstrate that our approach surpasses existing frameworks with respect to humour, fluency, and coherence, alleviating the problems of humour loss and linguistic interference.
\end{itemize}

\section{Related Work}
\subsection{Humour Theory} 
The incongruity theory \cite{raskin1979semantic} believes that the key to humour is the incongruity between readers' expectations and the ending of one story \cite{amir2016modelling}. Toplyn \cite{toplyn2014comedy} further proposes the monologue joke generation theory, which defines the structure of a joke as the topic, angle and punchline. There are currently some studies that incorporate humour theory into natural language processing for humour generation \cite{chen2023prompt} and humour recognition \cite{alnajjar2022laugh,kenneth2024two}. According to this theory, we will explore how to translate the jokes across different languages.

\subsection{Translation for LLMs} 
Extensive research has been conducted to evaluate the translation capabilities of LLMs. Some people study issues specific to LLMs, including the selection of prompt templates \cite{jiao2023chatgpt,zhang2023prompting} and In-Context Learning \cite{vilar2022prompting}. Other researchers investigate translation across diverse scenarios, such as low-resource translation \cite{jiao2023chatgpt}, document-level \cite{hendy2023good,karpinska2023large} and Multilingual machine translation \cite{zhu2023multilingual}. 

\begin{figure}
\centering
\includegraphics[width=\textwidth]{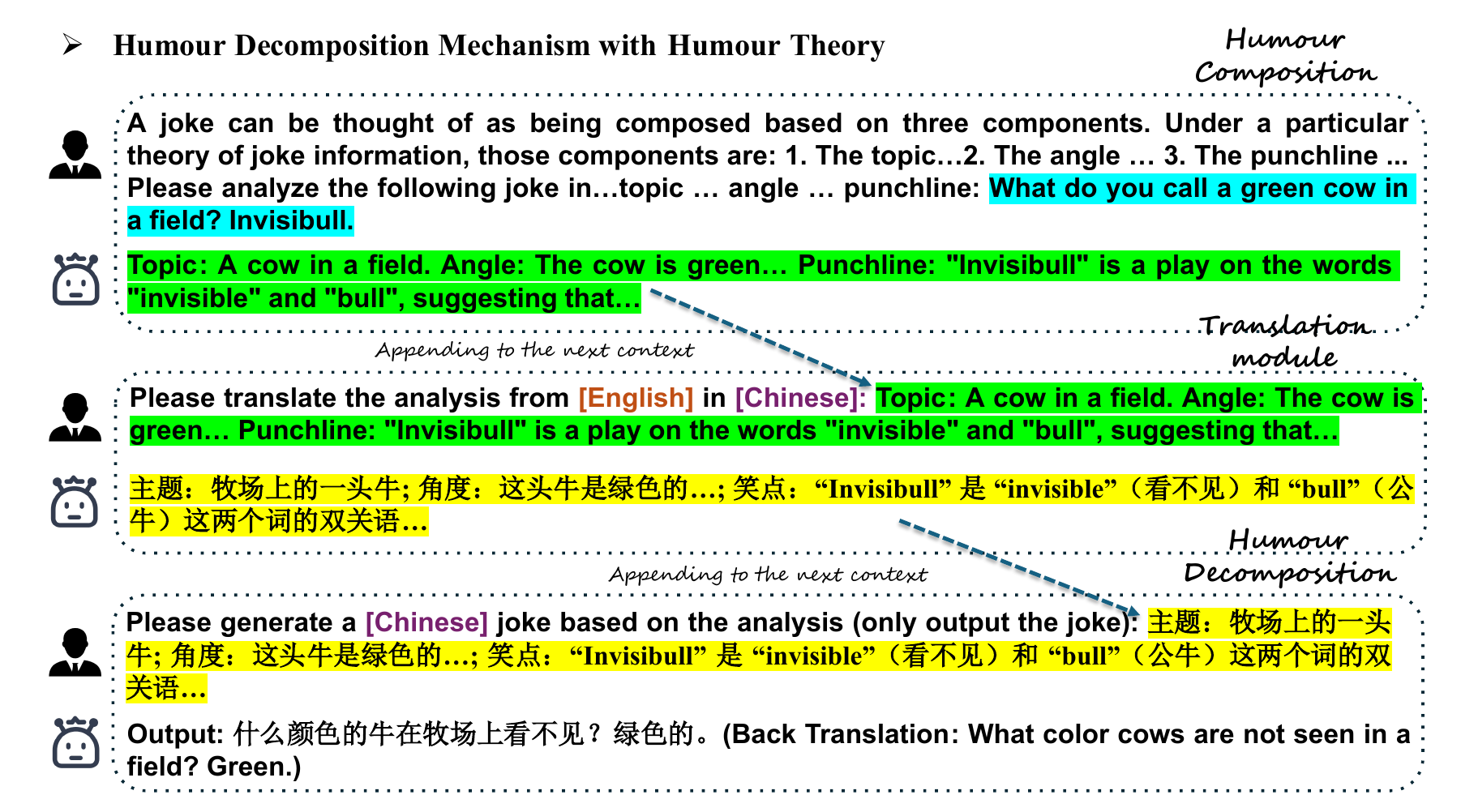}
\caption{The workflow of HDM.}
\label{flow}
\end{figure}

\subsection{Chain-of-Thought (CoT)} CoT prompting involves either providing instruction or a few chain-of-thought examples \cite{ji2024chain}. Recently, a series of studies \cite{ye2023explanation,zhou2022least} have proposed their respective prompting strategies, breaking down the entire task into smaller components and then systematically addressing, strategizing, and carrying out each of these components. With the improvement of model capabilities, some works \cite{zhou2022large,gao2023pal} treat the instruction as the “program” for searching, optimization, generating programs and bootstrapping the ability to perform successively more complex reasoning.

\section{Methodology}
Fig \ref{flow} illustrates an overview of the proposed Humour Decomposition Mechanism. Instead of directly asking LLMs for the final translation result, we hypothesise that the LLMs can analyze the latent humour interpretations and intrinsic knowledge before translating the jokes, and then generate the translated jokes based on this. We present two key contributions in this section.

\subsection{Humour Decomposition Mechanism}
We design three-step paradigm using Chain-of-Thought (CoT) prompting, which mimics the human thought process in solving complex reasoning tasks \cite{wei2022chain,wang2022self}, to enhance humour translation outcomes.

\subsubsection{Humour Decomposition} Humour decomposition is one of the important cores for HDM. Specifically, our approach initiates the LLM with a specific task of joke analysis. The request is formulated as follows:
\definecolor{shadecolor}{rgb}{0.92,0.92,0.92}
\begin{tcolorbox}
{You are a humour assistant. Please analyze the following joke: \sethlcolor{brightblue}\hl{[Given joke $\mathcal{L}_i$]}}
\end{tcolorbox}
Given a joke $\mathcal{L}_i$, we first claim the role of LLM in humour. Furthermore, we introduce an analysis process to generate the sequence of corresponding knowledge \emph{a}, which is organized into the final analysis $\mathcal{A}$. The formulation of our \emph{Humour Decomposition} method can be expressed as follows:
\begin{equation}
    \mathcal{A}_i=\arg \max p\left(\emph{a} \mid \mathcal{L}_i\right)
\end{equation}
where $\mathcal{L}_i$ and $\mathcal{A}_i$ denote the \emph{$i_{th}$} joke and its final analysis.

\subsubsection{Translation Module}
After achieving \emph{Humour Decomposition}, we use the \emph{Translation Module} to convert the source language analysis into the target language analysis. To illustrate, given the analysis $\mathcal{A}_i$ and the type of source language $\mathcal{S}$, we prompt the LLMs to translate $\mathcal{A}_i$ into target language $\mathcal{T}$, with the prompt defined as: 

\begin{tcolorbox}
    Please translate the analysis from \sethlcolor{orange}\hl{[source language $\mathcal{S}$]} into \sethlcolor{lightpurple}\hl{[target language $\mathcal{T}$]}: \sethlcolor{green}\hl{[text $\mathcal{A}_i$}]
\end{tcolorbox}

Formally, the translation is determined as:
\begin{equation}
    \mathcal{A}_i^{\prime}=\arg \max p\left(\emph{a}^{\prime} \mid \mathcal{A}_i,\mathcal{S},\mathcal{T},\right)
\end{equation}
where $\mathcal{A}_i^{\prime}$ represents the final translation of the analysis, generated from all potential translation results $\emph{a}^{\prime}$.

\subsubsection{Humour Composition}
Once the translation is generated, we further propose \emph{Humour Composition} to facilitate the generation of jokes. Given the translation version of the analysis, we design the prompt to make LLMs generate the joke of the target language. This is the structure of the prompt:
\begin{tcolorbox}[breakable]
{Please generate a \sethlcolor{lightpurple}\hl{[target language $\mathcal{T}$]} joke based on the analysis: \sethlcolor{yellow}\hl{[text $\mathcal{A}_i^{\prime}$}]}
\end{tcolorbox}
Formally, the humour composition can be defined as:
\begin{equation}
    \mathcal{F} =\arg \max p\left(\emph{f} \mid \mathcal{A}_i^{\prime},\mathcal{T}\right)
\end{equation}
where $\mathcal{F}$ is the final generation of the target language joke, generated from all potential generation results \emph{f}.

%

\subsection{Integrating Humour Theory}
In this section, we incorporate humour theory inspired by \cite{toplyn2014comedy} to enhance humour factors. The basic structure of the humorous text consists of the topic $\mathcal{X}$, angle $\mathcal{Y}$ and punchline $\mathcal{Z}$. The topic $\mathcal{X}$ is the news item that the joke is based on and the angle $\mathcal{Y}$ is the particular direction that the joke takes, while the punchline $\mathcal{Z}$ which is the surprise at the end of the joke. Therefore, the \emph{Humour Decomposition} module in HDM can be further improved as follows:
\definecolor{shadecolor}{rgb}{0.92,0.92,0.92}
\begin{tcolorbox}[breakable]

You are a humour assistant. A joke can be thought of as being composed based on three components. Under a particular theory of joke information, those components are:

1. The topic, which is the news item that the joke is based on. 

2. The angle, which is the particular direction that the joke takes. 

3. The punchline, which is the surprise at the end of the joke.

\end{tcolorbox}
Similarly, with \emph{Humour Decomposition}, we first claim the LLM’s role in humour. Then, we describe the components under the particular theory and give these components some details. Finally, we provide an instruction to format the model's outputs, which are defined as: 

\begin{tcolorbox}[breakable]
{Please analyze the following joke and provide the best explanation of what the topic is, what the angle is, and what the punchline is: \sethlcolor{brightblue}\hl{[Given joke $\mathcal{L}_i$]}}
\end{tcolorbox}
Formally, The improved formulation of the Humour Decomposition can be expressed as follows:
\begin{equation}
    \mathcal{A}_i=\arg \max p\left(\mathcal{X}_i,\mathcal{Y}_i,\mathcal{Z}_i \mid \mathcal{L}_i\right)
\end{equation}
where $\mathcal{A}_i$ denotes the analysis of the \emph{$i_{th}$} joke, including the con-cat of topic $\mathcal{X}_i$, angle $\mathcal{Y}_i$ and punchline $\mathcal{Z}_i$. 

HDM leverages the advanced generative capabilities of LLMs \cite{hagos2024recent} to reconstruct humour translation, overcoming the limitations of traditional translation methods, which are often constrained by linear word or phrase arrangements and linguistic interference, to improve the fluency and coherence of jokes. Additionally, the integration of humour theory defines the general structure of joke composition within the prompts, enabling the large language model to better comprehend background and punchline information. It theoretically enhances the LLM's ability to generate more humorous jokes, and we will also be demonstrated in our experiments.

\section{Experiments}

\subsection{Experimental Setup}
We select four representative state-of-the-art LLMs from the Chatbot Arena Leaderboard \cite{zheng2023judging} as backbone references for our study: Gemini1.5-Pro \cite{team2024gemini}, Yi-Large \cite{ai2024yi}, GPT3.5-Turbo and GPT4-Turbo. We set the default hyperparameters of these models in their playgrounds. Additionally, we use DUAL-REFLECT \cite{chen-etal-2024-dual} and MAPS \cite{he2024exploring}, which are the state-of-the-art translation approaches, as our baselines. Given budget constraints, we randomly select 500 samples on the Short Jokes Dataset \footnote{https://www.kaggle.com/datasets/thedevastator/short-jokes-dataset} for experiments. Finally, we evaluate their performance by using automatic metrics. 

\subsection{Metrics}
Since our approach specializes in humorous translation tasks, traditional automatic evaluation methods, such as COMET \cite{rei2020comet} and BLEURT \cite{sellam-etal-2020-bleurt}, have difficulty evaluating elements like humour. Therefore, inspired by \cite{kocmi2023large}, we evaluate the final results by using GEMBA which is a GPT4-based metric for generation quality. We choose the open area no-reference metrics GEMBA-SQM and GEMBA-STARS for their superior performance in \cite{kocmi2023large}. Specifically, GEMBA-SQM evaluates scalar quality metrics by dividing the assessment results into several stages, where 0 and 100 represent the lowest and highest scores, respectively. GEMBA-STARS is a classification task based on a one-to-five star ranking, which is a style often used when users are asked to review various services or products \cite{kocmi2023large}. In this section, SQM-H, SQM-F and SQM-C represent GEMBA-SQM metrics and STAR-H, STAR-F and STAR-C represent GEMBA-STARS metrics in humour, fluency and coherence.

To adapt to the evaluation of humour translation in linguistic interference and humour factor, we modify the original translation prompts and use the keywords of humour, coherence and fluency based on \cite{chen2024talk}. We report the performance by averaging the results over three runs in each type of experiment. Additionally, some answers are observed occasionally fall outside these ranges because of the LLM's hallucination \cite{kocmi2023large}. For example, instead of providing predicted scores, the model occasionally outputs explanations as results. Therefore, we omit the invalid responses and retain only the valid results in this research.



\subsection{Main Results}
As shown in Table \ref{table2}, HDM outperforms all baselines in terms of humour, fluency, and coherence in automatic metrics. This is particularly evident in the translation from English to Chinese in GPT4-Turbo, where the degree of humour improves by an average of 11.2\%. These results show that HDM can go beyond the other state-of-the-art translation methods, both enhancing the humour of translated text in humour translation and alleviating the problem of linguistic interference.


\begin{table}[t]
\centering

\begin{tabular}{l|cccccc}
\toprule
 Method & SQM-H & STAR-H & SQM-F & STAR-F & SQM-C    & STAR-C\\
\midrule
Gemini1.5-Pro & 49.82 & 2.53 & 96.74  &4.81 &89.30 &4.50 \\ \hspace{2em} + DUAL \cite{chen-etal-2024-dual}& 50.86 & 2.69 & 92.98 & 4.46 & 84.74 & 4.18 \\
\hspace{2em} + MAPS \cite{he2024exploring} & 57.98 & 3.01 & 96.35 & 4.74 & 89.95 & 4.48\\
\hspace{2em} + HDM & \textbf{63.80} & \textbf{3.19} &\textbf{98.54} &\textbf{4.93} &\textbf{94.27}    &\textbf{4.74} \\ \midrule
Yi-Large &53.40  &2.57 & 95.37&4.76 &86.58  &4.42 \\
\hspace{2em} + DUAL \cite{chen-etal-2024-dual} & 56.34 & 2.85 & 94.30  & 4.63& 87.01 & 4.34 \\
\hspace{2em} + MAPS \cite{he2024exploring} & 58.08 & 2.94 & 95.24 & 4.67 & 87.09 & 4.36\\
\hspace{2em} + HDM & \textbf{67.99} &\textbf{3.22} & \textbf{98.99}&\textbf{4.95} &\textbf{95.56}  &\textbf{4.85}\\ \midrule
GPT3.5-Turbo &50.03  &2.52 &94.33   &4.72 &86.83   &4.41 \\ 
\hspace{2em} + DUAL \cite{chen-etal-2024-dual} & 54.63 & 2.77 & 92.02 & 4.48 & 83.42 & 4.16\\
\hspace{2em} + MAPS \cite{he2024exploring} & 57.66 & 2.87 & 94.58 & 4.59 & 85.90 & 4.31\\
\hspace{2em} + HDM &\textbf{61.73} & \textbf{3.05} &\textbf{96.07} &\textbf{4.80} &\textbf{88.75}    &\textbf{4.49} \\ \midrule
GPT4-Turbo &53.20 &2.58 &94.95  &4.76 &87.70  &4.67\\
\hspace{2em} + DUAL \cite{chen-etal-2024-dual} & 58.33 & 2.95 & 91.60  & 4.43 & 83.30 & 4.13\\
\hspace{2em} + MAPS \cite{he2024exploring} &59.34 & 3.02 & 95.12 & 4.68 & 88.62 & 4.45 \\
\hspace{2em} + HDM & \textbf{70.54}  & \textbf{3.45} & \textbf{99.45} &\textbf{4.99} & \textbf{97.73}  &\textbf{4.96}\\ 
\bottomrule
\end{tabular}
\caption{Main results of the automatic metrics GEMBA-SQM and GEMBA-STARS in humour, fluency and coherence for translating from English to Chinese on the Short Joke Dataset. Both higher evaluation metrics indicate better performance.}
\label{table2}
\end{table}

\section{Analysis}
\subsection{Generality Analysis of HDM.}
To further investigate the generality of our work, we verify the generality of HDM from three perspectives \footnote{Given budget constraints, we have randomly selected 100 samples in each dataset and language.}. MAPS \cite{he2024exploring} is selected as the baseline for the generality analysis based on the comprehensive metrics evaluated in the experiment:
\subsubsection{Does HDM work well on other datasets?}
We conduct experiments on other datasets, namely the Question-Answer Jokes dataset \cite{QAdataset} and SemEval 2021 \cite{garcia-diaz-valencia-garcia-2021-umuteam}. Table \ref{otherdataset} shows that HDM can obtain better performance across all LLMs and metrics in different datasets, achieving improvements of at least 1.84\% in humour, 1.7\% in fluency and 2.15\% in coherence.

\begin{table}[t]
\setlength{\tabcolsep}{7pt}
\centering

\begin{tabular}{ccccccc}
\hline
\multirow{2}{*}{ Model } & \multicolumn{2}{c}{SQM-H} & \multicolumn{2}{c}{SQM-F} & \multicolumn{2}{c}{SQM-C}\\
& base & ours & base & ours & base & ours \\
\hline
    \rowcolor{gray!20}\multicolumn{7}{c}{Question-Answer Joke}\\
    Gemini1.5-Pro&60.00&\textbf{64.02}&97.67&\textbf{99.53}&85.29&\textbf{90.63}\\
    Yi-Large & 61.10 & \textbf{67.30} &96.00 & \textbf{99.00} &82.30 & \textbf{93.00} \\
    GPT3.5-Turbo  & 62.30 & \textbf{64.14} &95.45 & \textbf{97.37} &82.12 & \textbf{87.68} \\
    GPT4-Turbo & 64.70 & \textbf{68.70} &96.40 & \textbf{99.10} &87.37 &\textbf{95.05}\\
\hline
    \rowcolor{gray!20}\multicolumn{7}{c}{SemEval-2021}\\
    Gemini1.5-Pro&61.20&\textbf{64.60}&97.90&\textbf{99.00}&90.95&\textbf{93.10}\\
    Yi-Large & 57.90 & \textbf{67.50} &96.50 & \textbf{99.20} &92.85 & \textbf{95.35} \\
    GPT3.5-Turbo  & 56.30 & \textbf{66.06} &96.80 & \textbf{98.50} &88.05 & \textbf{93.35} \\
    GPT4-Turbo & 59.90 & \textbf{70.10} &96.70&\textbf{99.10} &91.70 &\textbf{97.20} \\
\hline
\end{tabular}

\caption{Generality analysis of automatic metric in translating from English to Chinese in different Datasets.}
\label{otherdataset}
\end{table}

\begin{table}[t]
\setlength{\tabcolsep}{7pt}
\centering

\begin{tabular}{ccccccc}
\hline
\small
\multirow{2}{*}{ Model } & \multicolumn{2}{c}{SQM-H} & \multicolumn{2}{c}{SQM-F} & \multicolumn{2}{c}{SQM-C}\\
& base & ours & base & ours & base & ours \\
\hline
    \rowcolor{gray!20}\multicolumn{7}{c}{EN=$>$SP}\\
 
Gemini1.5-Pro&59.50&\textbf{64.70}&94.00&\textbf{97.90}&87.20&\textbf{91.60}\\
    Yi-Large & 58.25 & \textbf{68.35} &\textbf{96.50} &96.30  &89.55 & \textbf{91.15} \\
    GPT3.5-Turbo  & 57.90 & \textbf{68.20} &95.70 & \textbf{97.00} &88.90 &\textbf{89.48}  \\
    GPT4-Turbo & 61.40 & \textbf{69.80} &95.53&\textbf{98.88} &89.50 &\textbf{95.50} \\
\hline
    \rowcolor{gray!20}\multicolumn{7}{c}{EN=$>$GE}\\
    Gemini1.5-Pro & 62.80 &\textbf{65.20}&95.10&\textbf{95.90}&89.00&\textbf{89.80}\\
    Yi-Large & 61.80 & \textbf{64.55} &94.25 & \textbf{97.50} &87.40 &\textbf{90.40}  \\
    GPT3.5-Turbo  & 61.80 & \textbf{65.30} &92.90&\textbf{97.30} &85.35 &\textbf{87.50}  \\
    GPT4-Turbo & 61.30 &\textbf{68.50} &95.90&\textbf{98.00} &88.70 &\textbf{89.85} \\
\hline
\end{tabular}

\caption{Generality analysis of automatic metric in different languages. \textit{SP} represents Spanish and \textit{GE} represents German.}
\label{otherlanguage}
\end{table}

\begin{table}[t]
\setlength{\tabcolsep}{6pt}
\centering

\begin{tabular}{ccccccc}
\hline
\multirow{2}{*}{ Model } & \multicolumn{2}{c}{SQM-H} & \multicolumn{2}{c}{SQM-F} & \multicolumn{2}{c}{SQM-C}\\
& base & ours & base & ours & base & ours \\
\hline
     Marco-o1-7B & 50.45 & \textbf{54.40} & 91.60 & \textbf{93.75} & \textbf{84.80} & 84.00 \\
    Llama3.1-8B-Instruct & 49.25 & \textbf{57.20} &\textbf{93.00} & 92.60 & \textbf{84.20}&81.70\\
    Qwen2.5-0.5B-Instruct&\textbf{39.80}&28.70&86.40&\textbf{89.10}&74.30&\textbf{80.50}\\
    Qwen2.5-7B-Instruct&51.40&\textbf{60.50}&92.50&\textbf{94.10}&85.60&\textbf{85.95}\\
    Qwen2.5-14B-Instruct & 51.60 & \textbf{61.80} & 91.40 & \textbf{97.40} & 86.00 & \textbf{91.35} \\
\hline
\end{tabular}

\caption{Generality analysis of automatic metric in translating from English to Chinese in different open-source models.}
\label{opensource}
\end{table}

\subsubsection{Does HDM work well on other language?}
To better assess the model's generalization capabilities, we conduct the experiments in different languages, including Spanish and German. As shown in Table \ref{otherlanguage}, the experimental results demonstrate that HDM consistently performs significantly well across these languages, for instance,  with improvements of 2.75\% in humour, 3.25\% in fluency, and 3\% in coherence in Yi-Large when translating from English to German. Those further demonstrate the effectiveness and broad applicability of HDM.

\subsubsection{Does HDM work well on open-source models?}
We also evaluate the performance of our methods using open-source models. Table \ref{opensource} presents the results for humour, fluency, and coherence on the Short Joke dataset, using mainstream open-source models: Qwen2.5-(0.5B, 7B, 14B) \cite{yang2024qwen2}, Llama3.1-8B-Instruct \footnote{https://huggingface.co/meta-llama/Meta-Llama-3-8B-Instruct} and Marco-o1-7B \cite{zhao2024marco}. We observe a significant decline in the evaluation scores for LLMs with smaller parameter sizes, especially for SQM-H. This may be attributed to their limited capacity to capture the intrinsic humorous semantics of the source text during the humour decomposition and humour composition, thereby impairing the humour quality of the final output.

\subsection{Prompt Selection}
We also validate the robustness of the zero-shot Humour Decomposition Mechanism against the different humour translation prompting.

\begin{figure}[ht]
  \centering
  \begin{subfigure}[b]{0.23\textwidth}
    \includegraphics[width=\textwidth]{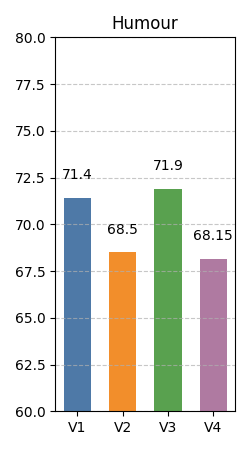}
    \label{fig:sub1}
  \end{subfigure}
  \hspace{-0.01\textwidth} 
  \begin{subfigure}[b]{0.23\textwidth}
    \includegraphics[width=\textwidth]{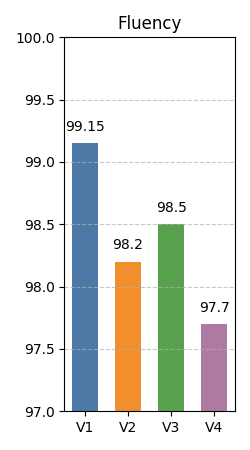}
    \label{fig:sub2}
  \end{subfigure}
  \hspace{-0.01\textwidth} 
  \begin{subfigure}[b]{0.23\textwidth}
    \includegraphics[width=\textwidth]{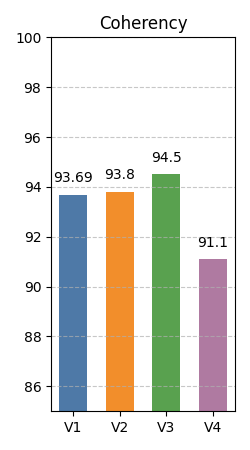}
    \label{fig:sub2}
  \end{subfigure}
  \caption{Performance comparisons of four various prompts of HDM in humour, fluency and coherency, marked by V1, V2, V3 and V4. The y-axis is the score on the GEMBA-SQM.}
  \label{prompt}
\end{figure}

Figure \ref{prompt} illustrates the performance of four different prompts in HDM by using GPT4-Turbo. The experimental findings reveal that despite fluctuations in GEMBA-SQM evaluation of reasoning across different prompts, all humour translation prompts consistently enhance performance compared to the traditional CoT approach. This further demonstrates the generalizability and effectiveness of the method, rather than its reliance on specific prompts.

\subsection{Ablation Study}
This analysis aims to investigate the effects of the results on Humour Theory and the Humour Decomposition Mechanism. We randomly select 100 samples to conduct the ablation study, as shown in Table \ref{ablation}, where:
\begin{itemize}
    \item ``-HT'' denotes removing the part of humour theory. Our approach will only use the analyzes for the intermediary stage.
    \item ``-HDM'' denotes removing the Humour Decomposition Mechanism. We directly input the prompt of decomposing humour to conduct the translation.
    \item ``base'' denotes both removing the Humour Decomposition Mechanism and humour theory.
\end{itemize}
From Table \ref{ablation} we observe that HDM demonstrates significant performance gains across all LLMs and evaluation metrics and plays a critical component of our approach, especially in humour. We attribute these improvements to CoT prompts, which help LLMs refine translated text by enhancing their parsing and reconstruction abilities.

Humour Theory (HT) further delivers some improvements after HDM. For example, Gemini1.5-Pro achieve gains of +3.3\%, +1.00\%, and +3.10\% in humour, fluency, and coherence, respectively. However, we find that the improvements are less pronounced after removing HDM compared to the baseline. In some cases, such as with GPT4, there are even declines. This indicates that HT works more effectively when combined with HDM, leading to better overall performance.

\begin{table}[t]
\setlength{\tabcolsep}{6pt}
\centering

\begin{tabular}{ccccccc}
\hline
\multirow{2}{*}{ Setting } & \multicolumn{3}{c}{EN=$>$ZH} & \multicolumn{3}{c}{EN=$>$SP} \\
& SQM-H & SQM-F & SQM-C & SQM-H & SQM-F & SQM-C \\

\hline
    \rowcolor{gray!20}\multicolumn{7}{c}{GPT4-Turbo}\\
    -  & \textbf{70.50} & \textbf{98.80} &\textbf{96.70} & \textbf{68.00} & \textbf{99.10} & \textbf{95.70} \\
    -HDM  & 54.60 & 95.65 &88.67 & 57.30 &97.00 & 89.10 \\
    -HT & 69.15 & 96.67 &91.77&67.10 &98.67 & 94.05\\
    base  & 51.60 & 93.30 &87.20& 55.20 & 96.67 & 88.80  \\
\hline
    \rowcolor{gray!20}\multicolumn{7}{c}{Gemini1.5-Pro}\\
    -  & \textbf{66.50} & \textbf{97.30} &\textbf{93.90}& \textbf{66.20} & \textbf{98.70} & \textbf{94.61} \\
    -HDM  & 57.40 & 94.00 &93.50& 60.80 & 98.30 & 89.40 \\
    -HT  & 63.20 & 96.30 &90.80& 65.80 & 98.40 & 92.00\\
    base  & 56.30 & 93.70 &87.70& 53.60 & 97.23 & 90.83 \\
\hline
\end{tabular}

\caption{Ablation results on Humour Decomposition Mechanism with various LLMs settings on Short Joke Dataset.}
\label{ablation}
\end{table}

\begin{figure*}[t]
\centering
\includegraphics[width=\textwidth]{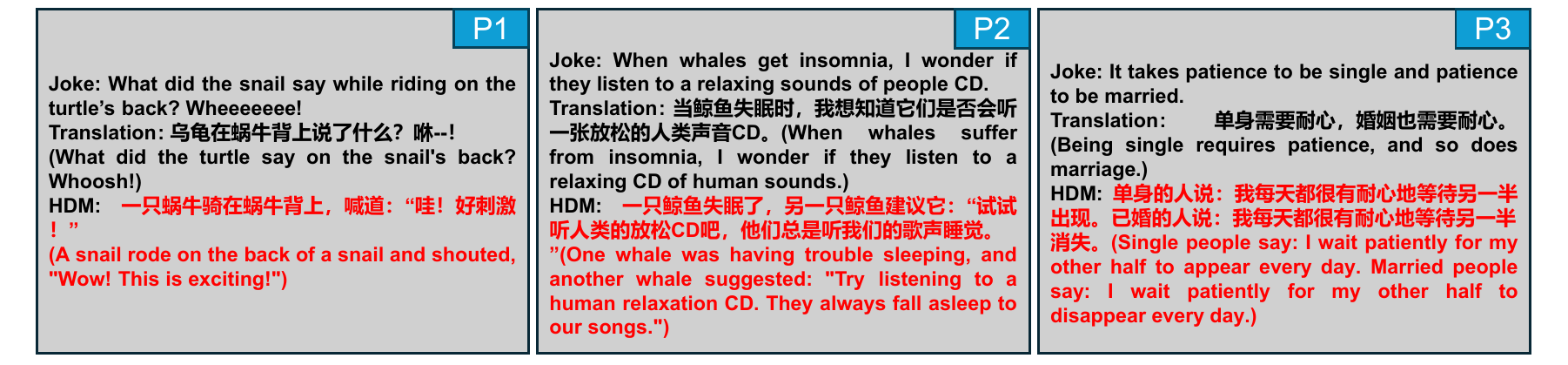} 
\caption{Some correct Chinese cases generated by HDM. We present the original jokes, traditional translations and their back translation and the results of HDM and their back translation.}
\label{good_case}
\end{figure*}
\subsection{Case Study and Error Analysis}
\label{case}
In this section, we present some correct examples generated by using HDM as shown in figure \ref{good_case} and make some analysis for some bad cases. For instance, the generated translation of $P_1$ describes the background sentence as ``the snail say while riding on the turtle's back'', while the snail shouting ``Wheeeeeee'' reflects the snail's feeling that the turtle is fast, which highlights the humorous effect. In the traditional translation, the onomatopoeia of ``Wheeeeeee'' is translated into ``Whoosh (back translation)'', while in HDM, the snail more intuitively reflects the language humour effect by saying ``Wow This is exciting! (back translation)''.  The jokes generated by using HDM are more informative and coherent than directly translated text, thus allowing people to better understand the humorous connotations of the texts.

In addition, there are still some samples that HDM is hard to address. One situation involves the judgment of the source language based on the pronunciation and shape of characters within the context of puns. For example, the joke is ``How do sheep in Mexico say Merry Christmas? Fleece Navidad!''. The punchline of this joke relies on the auditory similarity between ``Fleece'' and ``Feliz.'' By substituting ``Feliz'' with ``Fleece'' it creates a humorous image of sheep celebrating Christmas in their own way. In this case, HDM struggles to generate jokes that combine puns with cultural and linguistic elements.

\section{Limitations and Future Work}
While HDM shows the effectiveness in humour translation, there are some limitations worth noting: 
\begin{itemize}
    \item This study is limited to four languages: Chinese, English, Spanish, and German, with all datasets selected in English. Future research could extend both the data and the proposed method to additional translation directions.
    \item Although some studies have shown that GPT-4’s opinions significantly overlap with human reviewers \cite{liang2024can}, human evaluation remains the gold standard in assessing natural language generation quality \cite{clark2021all}. The absence of human evaluation in our current metrics may introduce bias into the final results. In future work, we plan to incorporate human judgments to enhance the robustness and validity of the evaluation.
    \item  Although Section \ref{case} provides both correct examples and error cases of HDM in humour translation, its applicability to different types of humour has not yet been systematically evaluated. Given the diversity of humour across regions and cultural contexts, further in-depth analysis is required to assess the method’s generalizability in broader translational scenarios.
\end{itemize}
Nevertheless, HDM provides meaningful insights into tackling the complex problems of humour loss and linguistic interference in humour translation, facilitating the effective transfer of humour across different languages.

\section{Conclusion}

In this paper, we introduce a novel approach named Humour Decomposition Mechanism (HDM) for humour translation. Specifically, HDM consists of \textit{humour decomposition}, \textit{translation module} and \textit{humour composition}, which creates a three-step paradigm of mining intrinsic knowledge of jokes, translating the intrinsic knowledge and then composing the jokes based on the translation. Moreover, we integrate humour theory into HDM to boost performance further. Experimental results in automatic evaluation reveal our method can attain promising performance in humour translation. In the future, we will explore the methods for incorporating human review in HDM to further improve the quality of humour translation.

\bibliographystyle{splncs04}
\bibliography{mybibliography}
%






\end{document}